\documentclass[10pt,twocolumn,letterpaper]{article}

\usepackage{cvpr}
\usepackage{times}
\usepackage{epsfig}
\usepackage{graphicx}
\usepackage{amsmath}
\usepackage{amssymb}

\usepackage[breaklinks=true,bookmarks=false]{hyperref}

\usepackage[table]{xcolor}
\usepackage{booktabs}
\usepackage{multirow}
\def\fps@figure{htp}
\def\fps@table{htp}

\newcommand{\bfig}{\begin{figure}}
\newcommand{\efig}{\end{figure}}

\newcommand{\benum}{\begin{enumerate}}
\newcommand{\eenum}{\end{enumerate}}

\newcommand{\ba}{\begin{eqnarray}}
\newcommand{\ea}{\end{eqnarray}}
\newcommand{\unit}[1]{\mbox{$\rm \,#1$}}

\cvprfinalcopy 

\begin{document}

\title{Tuning Modular Networks with Weighted Losses for Hand-Eye Coordination}
\author{Fangyi Zhang, J\"urgen Leitner, Michael Milford, Peter I.~Corke
  \thanks{FZ, JL, MM, PIC are with the Australian Centre for Robotic Vision (ACRV), Queensland University of Technology (QUT), Brisbane, Australia. {\tt\small fangyi.zhang@hdr.qut.edu.au}}
  \thanks{This research was conducted by the Australian Research Council Centre of Excellence for Robotic Vision (project number CE140100016). Additional computational resources and services were provided by the HPC and Research Support Group at QUT.}
}

\maketitle

\begin{abstract}
This paper introduces an end-to-end fine-tuning method to improve hand-eye coordination in modular deep visuo-motor policies (modular networks) where each module is trained independently. Benefiting from weighted losses, the fine-tuning method significantly improves the performance of the policies for a robotic planar reaching task.
\end{abstract}

\section{Introduction}
\label{sec:intro}


Recent work has demonstrated robotic tasks based directly on real image data using deep learning, for example robotic grasping~\cite{levine2016learning}.
However these methods require large-scale real-world datasets, which are expensive, slow to acquire and limit the general applicability of the approach.

\begin{figure}[tpb!]
\vspace{1.5mm}
\begin{center}
\includegraphics[width=1\columnwidth]{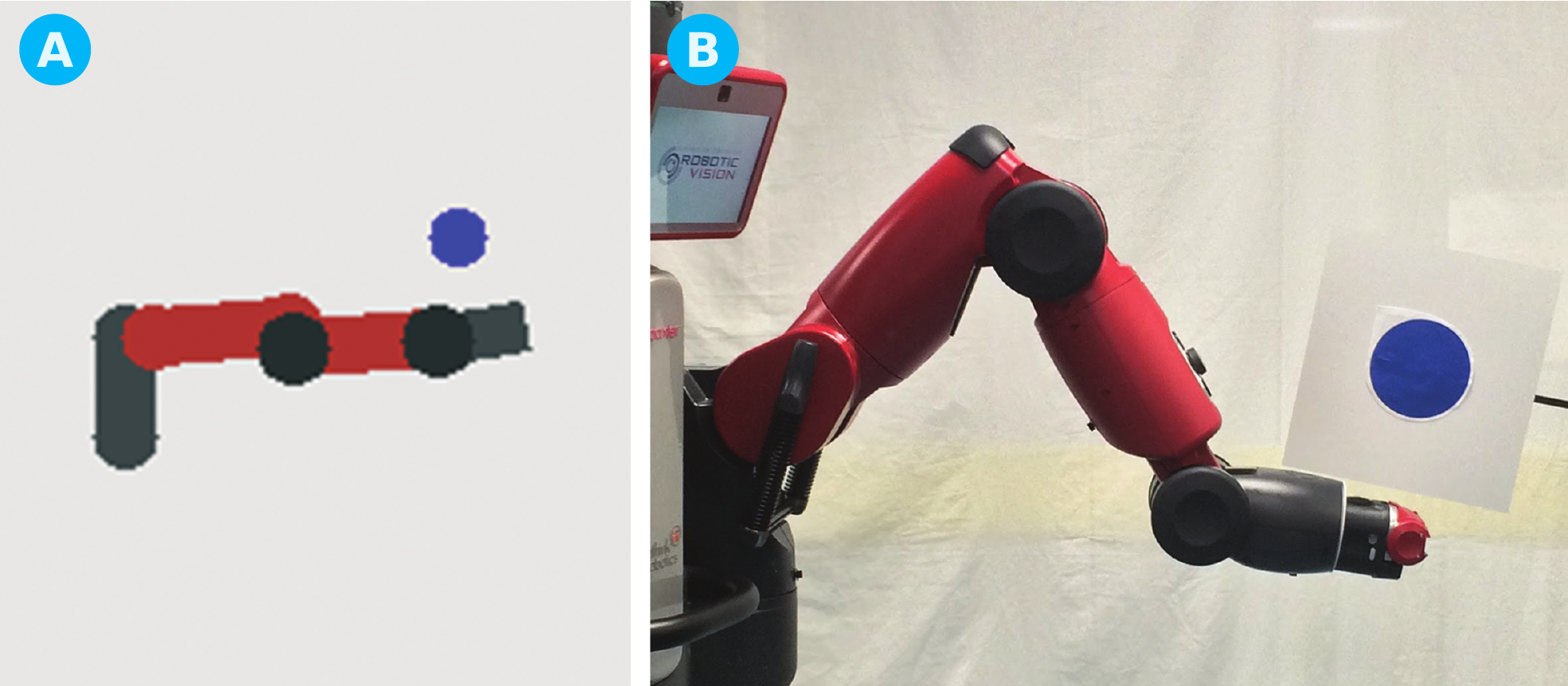}
\end{center}
\vspace{-2.5mm}
\caption{%
A technique to improve hand-eye coordination for better performance when transferring deep visuo-motor policies for a planar reaching task from simulated (A) to real environments (B).
 }
 
\label{fig:sim_to_real}
\vspace{-0.1cm}
\end{figure}

To reduce the cost of real dataset collection, we used simulation to learn robotic planar reaching skills  using the DeepMind DQN~\cite{mnih2015human}. The DQN showed impressive results in simulation, but exhibited brittleness when transferred to a real robot and camera~\cite{zhang2015towards}. By introducing a bottleneck to separate the DQN into perception and control modules for independent training, the skills learned in simulation (Fig.~\ref{fig:sim_to_real}A) were easily adapted to real scenarios (Fig.~\ref{fig:sim_to_real}B) by using just 1418 real-world images~\cite{zhang2017iros}.

However, there is still a performance drop compared to the control module network with ideal perception. 
To reduce the performance drop, we propose fine-tuning the combined network to improve hand-eye coordination. Preliminary studies show that a naive fine-tuning using Q-learning does not give the desired result~\cite{zhang2017iros}. 
To tackle the problem, we introduce a novel end-to-end fine-tuning method using weighted losses in this work, which significantly improved the performance of the combined network.

\section{Methodology}
\label{sec:methodology}

\begin{figure}[tb!]
\begin{center}
\includegraphics[width=1\columnwidth]{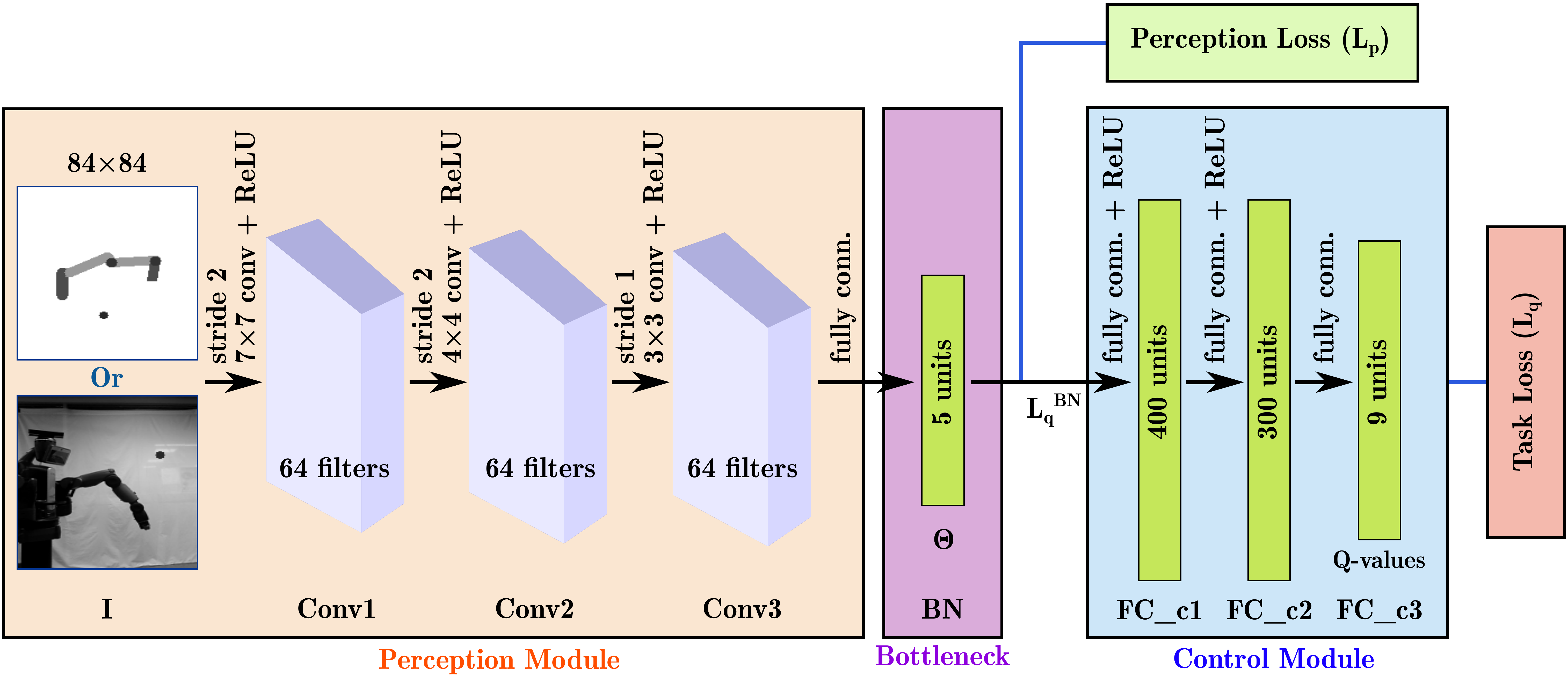}
\end{center}
\vspace{-0.25cm}
\caption{A modular neural network is used to predict Q-values given some raw pixel inputs. It is composed of perception and control modules. The perception module which consists of three convolutional layers and a FC layer, extracts the physically relevant information ($\mathbf{\Theta}$ in the bottleneck) from a single image. The control module predicts action Q-values given $\mathbf{\Theta}$. The action with a maximum Q-value is executed. The architecture is similar to that in~\cite{zhang2017iros}, but has an additional end-to-end fine-tuning process using weighted perception and task losses. Note that the values in $\mathbf{\Theta}$ are normalized to the interval $[0,1]$.}
\label{fig:net_architecture}
\vspace{-0.25cm}
\end{figure}

We consider the planar reaching task, which is defined as controlling a 3 DoF robot arm (Baxter robot's left arm) so that in operational space its end-effector position $\mathbf{x} \in \mathbb{R}^2$ moves to the position of the target  $\mathbf{x}^*$ in a vertical plane (ignoring orientation). The reaching controller adjusts the robot configuration (joint angles $\mathbf{q} \in \mathbb{R}^3$) to minimize the error between the robot's current and target position, i.e., $\left \| \mathbf{x} - \mathbf{x}^*\right \| $. At each time step 1 of 9 possible actions $a \in \mathbf{a}$ is chosen to change the robot configuration: 3 per joint -- increasing or decreasing by a constant amount (0.04\unit{rad}) or leaving it unchanged. An agent is required to learn to reach using only raw-pixel visual inputs $I$ from a monocular camera and their accompanying rewards $r$.

The network has the same architecture and training method to~\cite{zhang2017iros}, but with an additional end-to-end fine-tuning using weighted losses, as shown in Fig.~\ref{fig:net_architecture}. The perception network is first trained to estimate the scene configuration $\mathbf{\Theta} = [\mathbf{x^*} \, \mathbf{q}] \in \mathbb{R}^5$ from a raw-pixel image $I$ using the quadratic loss function
\begin{equation*}
\label{equ:perception_cost}
L_p=\frac{1}{2m} \sum_{j=1}^{m} \left \| y(I^j) - \mathbf{\Theta}^j \right \|^2,
\end{equation*}
where $y(I^j)$ is the prediction of $\mathbf{\Theta}^j$ for $I^j$; $m$ is the number of samples.
The control network is trained using K-GPS~\cite{zhang2017iros} where network weights are updated using the Bellman equation which is equivalent to the loss function 
\begin{equation*}
\label{equ:perception_cost}
L_q=\frac{1}{2m} \sum_{j=1}^{m} \left \| Q(\mathbf{\Theta}_t^j,a_t^j) - (r_{t}^j + \gamma \max_{a_{t+1}^j} Q(\mathbf{\Theta}_{t+1}^j,a_{t+1}^j)) \right \|^2,
\end{equation*}
where $Q(\mathbf{\Theta}_t^j,a_t^j)$ is the sum of future expected rewards $\sum_{k=0}^{\infty} \gamma^{k} r_{t+k}^j$ when taking action $a_t^j$ in state $\mathbf{\Theta}_t^j$. $\gamma$ is a discount factor applied to future rewards.

After  separate training for perception and control individually, an end-to-end fine-tuning is conducted for the combined network (perception + control) using weighted task ($L_q$) and perception ($L_p$) losses. The control network is updated using only $L_q$, while the perception network is updated using the weighted loss 
\begin{equation*}
\label{equ:perception_cost}
L= \beta L_p + (1-\beta) L_q^{BN},
\end{equation*}
where $L_q^{BN}$ is a pseudo-loss which reflects the loss of $L_q$ in the bottleneck (BN); $\beta \in [0,1]$ is a balancing weight. From the backpropagation algorithm~\cite{lecun-88}, we can infer that $\delta_L = \beta \delta_{L_p} + (1-\beta) \delta_{L_q^{BN}} $, where $\delta_L$ is the gradients resulted by $L$; $\delta_{L_p}$ and $\delta_{L_q^{BN}}$ are the gradients resulting  respectively from $L_p$ and $L_q^{BN}$ (equivalent to that resulting from $L_q$ in the perception module).

\section{Experiments and Results}
\label{sec:experiments}

We evaluated the feasibility of the proposed approach using the metrics of Euclidean distance error $d$ (between the end-effector and target) and average accumulated reward $\bar{R}$ (a bigger accumulated reward means a faster and closer reaching to a target) in 400 simulated trials. 
For comparison, we evaluated three networks: \textbf{Initial}, \textbf{Fine-tuned} and \textbf{CR}. \textbf{Initial} is a combined network without end-to-end fine-tuning, which is labelled as EE2 in ~\cite{zhang2017iros}
 (comprising \textbf{FT75} and \textbf{CR}). 
\textbf{FT75} and \textbf{CR} are the selected perception and control modules which have the best performance individually. \textbf{Fine-tuned} is obtained by fine-tuning \textbf{Initial} using the proposed approach. \textbf{CR} works as a baseline indicating performance upper-limit.

In fine-tuning, $\beta = 0.8$, we used a learning rate between 0.01 and 0.001, a mini-batch size of 64 and 256 for task and perception losses respectively, and an exploration possibility of 0.1 for K-GPS. These parameters were empirically selected.  
To make sure that the perception module remembers the skills for both simulated and real scenarios, the 1418 real samples were also used to obtain $\delta_{L_p}$. Similar to \textbf{FT75}, 75\% samples in a mini-batch were from real scenarios, i.e., at each weight updating step, 192 extra real samples were used in addition to the 64 simulated samples in the mini-batch for $\delta_{L_q}$.

\begin{figure}[tb!]
\begin{center}
\includegraphics[width=1\columnwidth]{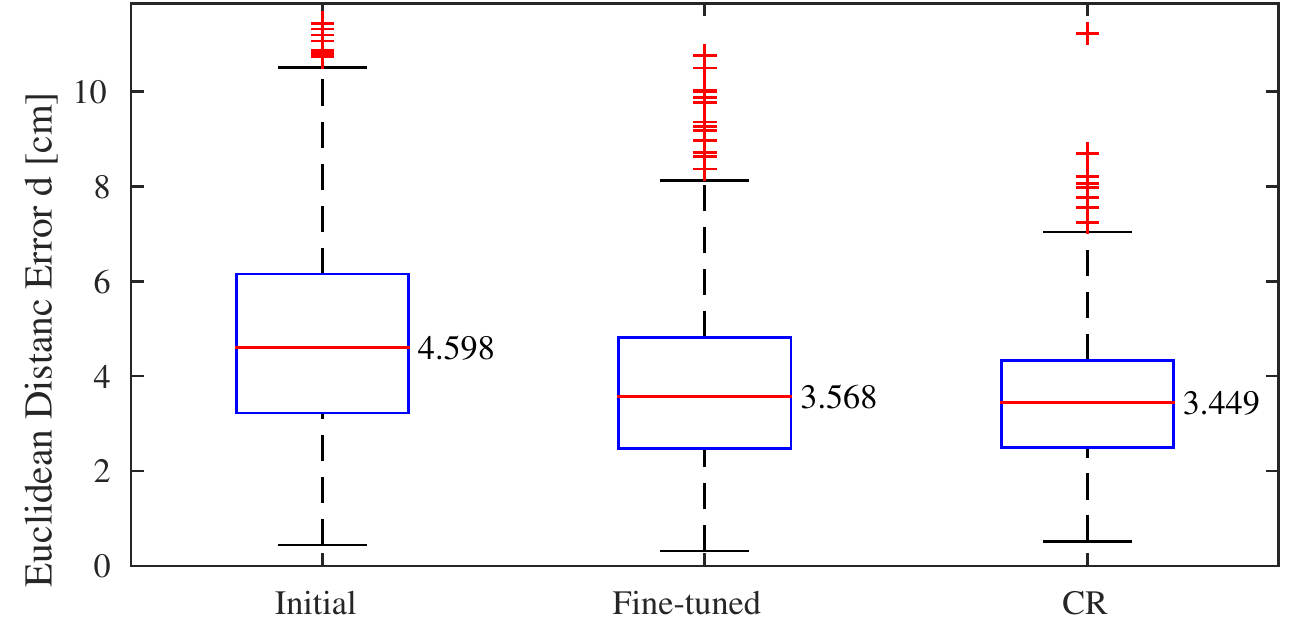}
\end{center}
\vspace{-0.35cm}
\caption{The box-plots of distance errors of different networks, with median values displayed. The crosses represent outliers.}
\label{fig:boxplot_d}
\vspace{-0.15cm}
\end{figure}

Results are summarized in Fig.~\ref{fig:boxplot_d} and Table~\ref{tab:e2e_experiments_results}. $d_{\mbox{med}}$ and $d_{Q3}$ are the median and third quartile of $d$. The error distance in pixels in the $84\times84$ input image is also listed.
We can see that \textbf{Fine-tuned} achieved a much better performance (22.4\% smaller $d_{\mbox{med}}$ and 96.2\% bigger $\bar{R}$) than \textbf{Initial}. 
The fine-tuned performance is even very close to that of the control module (\textbf{CR}) which controls the arm using ground-truth $\mathbf{\Theta}$ as sensing inputs. We also did the same evaluations in 20 real-world trials on Baxter, and achieved similar results. 



\begin{table}[tb!]
\caption{Planar Reaching Performance}
\label{tab:e2e_experiments_results}
\centering	
\renewcommand\arraystretch{1.1}
\renewcommand\tabcolsep{4pt}
\begin{tabular}{c | c | c | c | c | c }	
\toprule 

\multirow{2}{*}{Nets} & 

\multicolumn{2}{c|}{$d_{\mbox{med}}$} &
\multicolumn{2}{c|}{$d_{Q3}$} &
$\bar{R}$\\

\ & \bfseries [cm] & \bfseries [pixels] & \bfseries [cm] & \bfseries [pixels] & \bfseries [\textbackslash] \\ 

\hline 
{\bfseries Initial}  & 4.598 & 1.929 & 6.150 & 2.581 & 0.319 \\
{\bfseries Fine-tuned}  & 3.568 & 1.497 & 4.813 & 2.020 & 0.626 \\
\hline
{\bfseries CR}  & 3.449 & 1.447 & 4.330 & 1.817 & 0.761 \\





\bottomrule 
\end{tabular}
\vspace{-0.2cm}
\end{table}

The experimental results show the feasibility of the proposed fine-tuning approach. Improved hand-eye coordination in modular deep visuo-motor policies is possible due to fine-tuning with weighted losses. 
The adaptation to real scenarios can still be kept by presenting (a mix of simulated and) real samples to compute the perception loss.

{\footnotesize
\bibliographystyle{ieee}
\bibliography{deep_manipulation,rl,nc}

\begin{thebibliography}{1}\itemsep=-1pt

\bibitem{lecun-88}
Y.~LeCun.
\newblock A theoretical framework for back-propagation.
\newblock In D.~Touretzky, G.~Hinton, and T.~Sejnowski, editors, {\em
  Proceedings of the 1988 Connectionist Models Summer School}, pages 21--28,
  CMU, Pittsburgh, Pa, 1988. Morgan Kaufmann.

\bibitem{levine2016learning}
S.~Levine, P.~P. Sampedro, A.~Krizhevsky, and D.~Quillen.
\newblock Learning hand-eye coordination for robotic grasping with deep
  learning and large-scale data collection.
\newblock In {\em International Symposium on Experimental Robotics (ISER)},
  2016.

\bibitem{mnih2015human}
V.~Mnih, K.~Kavukcuoglu, D.~Silver, A.~A. Rusu, J.~Veness, M.~G. Bellemare,
  A.~Graves, M.~Riedmiller, A.~K. Fidjeland, G.~Ostrovski, et~al.
\newblock Human-level control through deep reinforcement learning.
\newblock {\em Nature}, 518(7540):529--533, 2015.

\bibitem{zhang2015towards}
F.~Zhang, J.~Leitner, M.~Milford, B.~Upcroft, and P.~Corke.
\newblock Towards vision-based deep reinforcement learning for robotic motion
  control.
\newblock In {\em Australasian Conference on Robotics and Automation (ACRA)},
  2015.

\bibitem{zhang2017iros}
F.~Zhang, J.~Leitner, B.~Upcroft, and P.~Corke.
\newblock Transferring vision-based robotic reaching skills from simulation to
  real world.
\newblock Technical report, 2017.

\end{thebibliography}
}

\end{document}